\documentclass[11pt]{article}
\usepackage{coling2020}
\usepackage{latexsym}
\usepackage{times}
\usepackage{url}
\usepackage{latexsym}
\usepackage{multirow}
\usepackage{amsmath}
\usepackage{adjustbox}
\usepackage{array}
\usepackage{subfig}
\usepackage{graphicx}
\usepackage{enumitem}
\usepackage{footnote}
\usepackage{booktabs}
\usepackage{xspace}
\usepackage{makecell}
\makesavenoteenv{tabular}
\makesavenoteenv{table}

\newcolumntype{L}[1]{>{\raggedright\let\newline\\\arraybackslash\hspace{0pt}}m{#1}}
\newcolumntype{C}[1]{>{\centering\let\newline\\\arraybackslash\hspace{0pt}}m{#1}}
\newcolumntype{R}[1]{>{\raggedleft\let\newline\\\arraybackslash\hspace{0pt}}m{#1}}

\newcommand{\secref}[2][]{Section#1~\ref{sec:#2}}

\newcommand{\tabref}[2][]{Table#1~\ref{tab:#2}}
\newcommand{\figref}[2][]{Figure#1~\ref{fig:#2}}

\newcommand{\lex}[1]{\textit{#1}\xspace}
\newcommand{\means}[1]{``#1''\xspace}

\newcommand{\dataset}[1]{\textsc{#1}\xspace}
\newcommand{\indolem}{\dataset{IndoLEM}}

\newcommand{\class}[1]{\texttt{#1}\xspace}

\newcommand{\model}[1]{\textsc{#1}\xspace}
\newcommand{\fasttext}{\texttt{fastText}}
\newcommand{\mbert}{\model{mBERT}}
\newcommand{\indobert}{\model{IndoBERT}}
\newcommand{\malaybert}{\model{MalayBERT}}
\newcommand{\sentfeat}{\ensuremath{\dagger}}

\colingfinalcopy 

\title{IndoLEM and IndoBERT: A Benchmark Dataset and Pre-trained Language Model for Indonesian NLP}

\author{Fajri Koto$^1$ \quad Afshin Rahimi$^2$ \qquad Jey Han Lau$^1$ \qquad Timothy Baldwin$^1$\\
	 $^1$The University of Melbourne \\
	 $^2$The University of Queensland \\
	\texttt{\small ffajri@student.unimelb.edu.au, afshinrahimi@gmail.com} \\
	\texttt{\small jeyhan.lau@gmail.com, tb@ldwin.net}
}

\date{}

\begin{document}

\maketitle

\begin{abstract}
  Although the Indonesian language is spoken by almost 200 million
  people and the 10th most-spoken language in the
  world,\footnote{\url{https://www.visualcapitalist.com/100-most-spoken-languages/}}
  it is under-represented in NLP research. Previous work on Indonesian
  has been hampered by a lack of annotated datasets, a sparsity of
  language resources, and a lack of resource standardization. In this
  work, we release the \indolem dataset comprising seven tasks for the
  Indonesian language, spanning morpho-syntax, semantics, and
  discourse. We additionally release \indobert, a new pre-trained
  language model for Indonesian, and evaluate it over \indolem, in
  addition to benchmarking it against existing resources.  Our
  experiments show that \indobert achieves state-of-the-art performance
  over most of the tasks in \indolem.
		
		
\end{abstract}
	
\section{Introduction}
\label{intro}
	
Despite there being over 200M first-language speakers of the Indonesian
language, the language is under-represented in NLP. We argue that there
are three root causes: a lack of annotated datasets, a sparsity of
language resources, and a lack of resource standardization. In English,
on the other hand, there are ever-increasing numbers of datasets for
different tasks
\cite{hermann2015teaching,luong2016ac,rajpurkar2018know,agirre2016semeval},
(pre-)trained models for language modelling and language understanding
tasks \cite{devlin2019bert,yang2019xlnet,radford2019language}, and
standardized tasks to benchmark research progress
\cite{wang2019glue,wang2019superglue,williams2018a}, all of which have
contributed to rapid progress in the field in recent years.
	
We attempt to redress this situation for Indonesian, as follows. First,
we introduce \indolem (``\underline{Indo}nesian \underline{L}anguage
\underline{E}valuation \underline{M}ontage''\footnote{Yes, guilty as
  charged, a slightly-forced backronym from \lex{lem}, which is
  Indonesian for \means{glue}, following the English benchmark naming
  trend (e.g.\ GLUE \cite{wang2019glue} and SuperGLUE
  \cite{wang2019superglue}).}), a comprehensive dataset encompassing
seven NLP tasks and eight sub-datasets, five of which are based on
previous work and three are novel to this work. As part of this, we
standardize data splits and evaluation metrics, to enhance
reproducibility and robust benchmarking.  These tasks are intended to
span a broad range of morpho-syntactic, semantic, and discourse analysis
competencies for Indonesian, to be able to benchmark progress in
Indonesian NLP. First, for morpho-syntax, we examine part-of-speech
(POS) tagging \cite{dinakaramani2014designing}, dependency parsing with
two Universal Dependency (UD) datasets, and two named entity recognition
(NER) tasks using public data. For semantics, we examine sentiment
analysis and single-document summarization. For discourse, we create two
Twitter-based document coherence tasks: Twitter response prediction (as
a multiple-choice task), and Twitter document thread ordering.
	
Second, we develop and release \indobert, a monolingual pre-trained BERT
language model for Indonesian \cite{devlin2019bert}. This is one of the
first monolingual BERT models for the Indonesian language, trained
following the best practice in the field.\footnote{Turns out we weren't
  the first to think to train a monolingual BERT model for Indonesian,
  or to name it IndoBERT, with (at least) two contemporaneous BERT
  models for Indonesian that are named ``IndoBERT'':
  \newcite{azhari2020indobert} and \newcite{wilie2020indonlu}.}
	
Our contributions in this paper are: (1) we release \indolem, which is
by far the most comprehensive NLP dataset for Indonesian, and intended
to provide a benchmark to catalyze further NLP research on the language;
(2) as part of \indolem, we develop two novel discourse tasks and
datasets; and (3) we follow best practice in developing and releasing
for general use \indobert, a BERT model for Indonesian, which we show to
be superior to existing pre-trained models based on \indolem. The
\indolem dataset, \indobert model, and all code associated with this
paper can be accessed at: \url{https://indolem.github.io}.

\section{Related Work}
\label{sec:related}
	
To comprehensively evaluate natural language understanding (NLU) methods
for English, collections of tools and corpora such as GLUE
\cite{wang2019glue} and SuperGLUE \cite{wang2019superglue} have been
proposed. Generally, such collections aim to benchmark models across
various NLP tasks covering a variety of corpus sizes, domains, and task
formulations. GLUE comprises nine language understanding tasks built on
existing public datasets,
while SuperGLUE is a set of eight tasks that is not only diverse in task
format but also includes low-resource settings. SuperGLUE is a more
challenging framework, and BERT models trail human performance by 20
points at the time of writing.
	
In the cross-lingual setting, XGLUE \cite{liang2020xglue} was introduced
as a benchmark dataset that covers nearly 20 languages. Unlike GLUE,
XGLUE includes language generation tasks such as question and headline
generation. One of the largest cross-lingual corpora is dependency
parsing provided by Universal
Dependencies.\footnote{\url{https://universaldependencies.org/}} It has
consistent annotation of 150 treebanks across 90 languages, constructed
through an open collaboration involving many contributors. Recently,
other cross-lingual benchmarks have been introduced, such as
\newcite{hu2020xtreme} and \newcite{lewis2020mlqa}. While these three
cross-lingual benchmarks contain some resources/datasets for Indonesian,
the coverage is low and data is limited.
	
Beyond the English and cross-lingual settings,
ChineseGLUE\footnote{\url{https://github.com/ChineseGLUE/ChineseGLUE}}
is a comprehensive NLU collection for Mandarin Chinese, covering eight
different tasks. For the Vietnamese language,
\newcite{nguyen2020phobert} gathered a dataset covering four tasks (NER,
POS tagging, dependency parsing, and language inference), and
empirically evaluated them against a monolingual BERT. Elsewhere, there
are individual efforts to maintain a systematic catalogue of tasks and
datasets, and state-of-the-art methods for each across multiple
languages,\footnote{\url{https://github.com/sebastianruder/NLP-progress}}
including one specifically for
Indonesian.\footnote{\url{https://github.com/kmkurn/id-nlp-resource}}
However, there is no comprehensive dataset for evaluating NLU systems in
the Indonesian language, a void which we seek to fill with \indolem.
	
\section{\indobert}
	
Transformers \cite{vaswani2017attention} have driven substantial
progress in NLP research based on pre-trained models in the last few
years. Although attention-based models are data- and GPU-hungry, the
full attention mechanisms and parallelism offered by the transformer are
highly compatible with the high levels of parallelism that GPU
computation offers, and have been shown to be highly effective at
capturing the syntax \cite{jawahar2019what} and sentence semantics of
text \cite{sun2019utilizing}.  In particular, transformer-based language
models
\cite{devlin2019bert,radford2018building,conneau2019cross,raffel2019exploring}
pre-trained on large volumes of text based on simple tasks such as
masked word prediction and sentence ordering prediction, have quickly
become ubiquitous in NLP and driven substantial empirical gains across
tasks including NER \cite{devlin2019bert}, POS tagging
\cite{devlin2019bert}, single document summarization \cite{liu2019text},
syntactic parsing \cite{kitaev-etal-2019-multilingual}, and discourse
analysis \cite{nie-etal-2019-dissent}. However, this effect has been
largely observed for high-resource languages such as English.
	
\indobert is a transformer-based model in the style of BERT
\cite{devlin2019bert}, but trained purely as a masked language model
trained using the Huggingface\footnote{\url{https://huggingface.co/}}
framework, following the default configuration for BERT-Base
(uncased). It has 12 hidden layers each of 768d, 12 attention heads, and
feed-forward hidden layers of 3,072d. We modify the Huggingface
framework to read a separate text stream for different document
blocks,\footnote{The existing implementation merges all documents into
  one text stream} and set the training to use 512 tokens per batch. We
train \indobert with 31,923-size Indonesian WordPiece vocabulary.
	
In total, we train \indobert over 220M words, aggregated from three main
sources: (1) Indonesian Wikipedia (74M words); (2) news articles from
Kompas,\footnote{\url{https://kompas.com}}
Tempo\footnote{\url{https://koran.tempo.co}} \cite{tala2003the}, and
Liputan6\footnote{\url{https://liputan6.com}} (55M words in total); and
(3) an Indonesian Web Corpus \cite{medved2019indonesian} (90M
words). After preprocessing the corpus into 512-token document blocks,
we obtain 1,067,581 train instances and 13,985 development instances
(without reduplication). In training, we use 4 Nvidia V100 GPUs (16GB
each) with a batch size of 128, learning rate of 1e-4, the Adam
optimizer, and a linear scheduler. We trained the model for 2.4M steps
(180 epochs) for a total of 2 calendar months,\footnote{We checkpointed
  the model at 1M and 2M steps, and found that 2M steps yielded a lower
  perplexity over the dev set.}  with the final perplexity over the
development set being 3.97 (similar to English BERT-base).
	
\section{\indolem: Tasks}
	
In this section, we present an overview of \indolem, in terms of the NLP
tasks and sub-datasets it includes. We group the tasks into three
categories: morpho-syntax/sequence labelling, semantics, and discourse
coherence. We summarize the sub-datasets include in \indolem in
\tabref{datastat}, in addition to detailing related work on the
respective tasks.
	
\begin{table}[t!]
  \begin{center}
        \begin{adjustbox}{max width=1\linewidth}
          \begin{tabular}{lrrrcc}
            \toprule
            \textbf{Data} & \textbf{\#train} & \textbf{\#dev} & \textbf{\#test} & \textbf{5-Fold} & \textbf{Evaluation} \\
            \midrule
            \multicolumn{5}{l}{\textbf{Morpho-syntax/Sequence Labelling Tasks}} \\
            POS Tagging* & 7,222 & 802 & 2,006 & Yes & Accuracy\\
            NER UI& 1,530 & 170 & 425 & No & micro-averaged F1\\
            NER UGM & 1,687 & 187 & 469 & No & micro-averaged F1\\
            UD-Indonesian GSD* & 4,477& 559 & 557 & No & UAS, LAS\\
            UD-Indonesian PUD (Corrected Version) & 700 & 100 & 200 & Yes & UAS, LAS\\
            \midrule
            \multicolumn{5}{l}{\textbf{Semantic Tasks}} \\
            Sentiment Analysis & 3,638 & 399 & 1,011 & Yes & F1 \\
            IndoSum* & 14,262 & 750 & 3,762 & Yes & ROUGE \\
            \midrule
            \multicolumn{5}{l}{\textbf{Coherency Tasks}} \\
            Next Tweet Prediction (NTP) & 5,681 & 811 & 1,890 & No & Accuracy\\
            Tweet Ordering & 5,327 & 760 & 1,521 & Yes & Rank Corr \\
					
            \bottomrule
          \end{tabular}
        \end{adjustbox}
      \end{center}
      \caption{Summary of datasets incorporated in \indobert. Datasets marked with `*' were already available with canonical splits.}
      \label{tab:datastat}
\end{table}

\subsection{Morpho-syntax and Sequence Labelling Tasks}
              
\textbf{Part-of-speech (POS) tagging}. The first Indonesian POS tagging
work was done over a 15K-token
dataset. \newcite{pisceldo2009probabilistic} defines 37 tags covering
five main POS tags: \textit{kata kerja} (verb), \textit{kata sifat}
(adjective), \textit{kata keterangan} (adverb), \textit{kata benda}
(noun), and \textit{kata tugas} (function words). They utilized news
domain and partial data from the PanLocalisation project
(``PANL10N''\footnote{\url{http://www.panl10n.net/}}). In total,
``PANL10N'' comprises 900K tokens, and was generated by
machine-translating an English POS-tagged dataset and noisily projecting
the POS tags from English to the Indonesian translations.
              
To create a larger and more reliable corpus,
\newcite{dinakaramani2014designing} published a manually-annotated
corpus of 260K tokens (10K sentences). The text was sourced from the
IDENTIC parallel corpus \cite{larasati2012identic}, which was translated
from data in the Penn Treebank corpus. The text is manually annotated
with 23 tags based on Indonesian tag definition of
\newcite{adriani2009statistical}.  For \indolem, we use the Indonesian
POS tagging dataset of \newcite{dinakaramani2014designing}, and 5-fold
partitioning of \newcite{kurniawan2018toward}.\footnote{We do not
  include POS data from the Universal Dependency project, as we found
  the data to contain many foreign borrowings (without any attempt to
  translate them into Indonesian), and some sentences to be poor
  translations (a point we return to in the context of error analysis of
  dependency parsing in \secref{results}).}
              
\textbf{Named entity recognition (NER)}. \newcite{budi2005named} was the
first study on named entity recognition for Indonesian, where roughly
2,000 sentences from a news portal were annotated with three NE classes:
\class{person}, \class{location}, and \class{organization}. In other work,
\newcite{luthfi2014building} utilized Wikipedia and DBPedia to
automatically generate an NER corpus, and trained a model with Stanford
CRF-NER \cite{finkel2005incorporating}. \newcite{rachman2017named}
studied LSTM performance over 480 tweets with the same three named
entity classes. None of these authors released the datasets used in the
research.
	
There are two publicly-available Indonesian NER datasets. The first, NER
UI, comprises 2,125 sentences obtained via an annotation assignment in
an NLP course at the University of Indonesia in 2016
\cite{nergithubgultom2017automatic}. The corpus has the same three named
entity classes as its predecessors \cite{budi2005named}. The second, NER
UGM, comprises 2,343 sentences from news articles, and was constructed
at the University of Gajah Mada \cite{nerugmFachri2014} based on five
named entity classes: \class{person}, \class{organization}, \class{location}, \class{time}, and
\class{quantity}.
	
\textbf{Dependency parsing}. \newcite{kamayani2011dependency} and
\newcite{green2012indonesian} pioneered dependency parsing for the
Indonesian language. \newcite{kamayani2011dependency} developed
language-specific dependency labels based on 20 sentences, adapted from
Stanford Dependencies \cite{marnee2010stanford}.
\newcite{green2012indonesian} annotated 100 sentences of IDENTIC without
dependency labels, and used an ensemble SVM model to build a parser.
Later, \newcite{rahman2017ensemble} conducted a comparative evaluation
over models trained using off-the-shelf tools such as MaltParser
\cite{nivre2005maltparser} on 2,098 annotated sentences from the news
domain.  However, this corpus is not publicly available.
	
The Universal Dependencies (UD)
project\footnote{\url{https://universaldependencies.org/}} has released
two different Indonesian corpora of relatively small size: (1) 5,593
sentences of UD-Indo-GSD
\cite{mcdonald2013universal};\footnote{\url{https://github.com/UniversalDependencies/UD_Indonesian-GSD}}
and (2) 1,000 sentences of UD-Indo-PUD
\cite{zeman2018conll}.\footnote{\url{https://github.com/UniversalDependencies/UD_Indonesian-PUD}}
\newcite{alfina2019gold} found that these corpora contain annotation
errors and did not deal adequately with Indonesian morphology. They released a
corrected version of UD-Indo-PUD by fixing annotations for
reduplicated-words, clitics, compound words, and noun phrases.
	
We include two UD-based dependency parsing datasets in \indolem: (1)
UD-Indo-GSD, and (2) the corrected version of UD-Indo-PUD. As our
reference dependency parser model, we use the BiAffine dependency parser
\cite{dozat2017deep}, which has been shown to achieve strong performance
for English.
	

\subsection{Semantic Tasks}
	
\textbf{Sentiment analysis}. There has been sentiment analysis for
Indonesian domains/data sources including presidential elections
\cite{ibrahim2015buzzer}, stock prices \cite{cakra2015stock}, Twitter
\cite{koto2017inset}, and movie reviews
\cite{nurdiansyah2018sentiment}. Most previous work, however, has used
non-public and low-resource datasets.
	
	
We include in \indolem an Indonesian sentiment analysis dataset based on
binary classification. In total, the data distribution is 3638/399/1011
sentences for train/development/test, respectively. The data was sourced
from Twitter \cite{koto2017inset} and hotel
reviews.\footnote{\url{https://github.com/annisanurulazhar/absa-playground/}}
The hotel review data is annotated at the aspect level, where one review
can have multiple polarities for different aspects. We simply count the
proportion of positive and negative polarity aspects, and label the
sentence based on the majority sentiment. We discard a review if there
is a tie in positive and negative aspects.

\textbf{Summarization}. From attention mechanisms
\cite{rush2015a,see2017get} to pre-trained language models
\cite{liu2019text,zhang2019pegasus}, recent summarization work on
English in terms of both extractive and abstractive methods has relied
on ever-larger datasets and data-hungry methods.
	
Indonesian (single document) text summarization research has inevitably
focused predominantly on extractive methods, based on small
datasets. \newcite{aris2012text} deployed a genetic algorithm over a
200-document summarization dataset, and \newcite{gunawan2017automatic}
performed unsupervised summarization over 3,075 news articles. As an
attempt to create a standardized corpus, \newcite{koto2016a} released a
300-document chat summarization dataset, and
\newcite{kurniawan2018indosum} released the \textit{IndoSum} 19K
document--summary dataset. At the time we carried out this work,\footnote{Noting that the soon-to-be-released Liputan6
  dataset \cite{Koto+:2020a-pre} will be substantially larger, but was
  not available when this research was carried out.} \textit{IndoSum} was the largest
Indonesian summarization corpus in the news domain, manually constructed
from CNN Indonesia\footnote{\url{https://www.cnnindonesia.com/}} and
Kumparan\footnote{\url{https://kumparan.com/}}
documents. \textit{IndoSum} is a
single-document summarization dataset where each article has one
abstractive summary. \newcite{kurniawan2018indosum} released
\textit{IndoSum} together with the \textsc{Oracle} --- a set of
extractive summaries generated automatically by maximizing ROUGE score
between sentences of the article and its abstractive summary.  We
include \textit{IndoSum} as the summarization dataset in \indolem, and
evaluate the performance of extractive summarization in this paper.
	

	
\subsection{Discourse Coherence Tasks}
	
	
We also introduce two tasks that measure the ability of models to
measure discourse coherence in Indonesian, based on message ordering in
Twitter threads, namely: (1) next tweet prediction; and (2) message
ordering. Utilizing tweets instead of edited text arguably makes the
task harder and allows us to assess the robustness of models.
	
First, we use the standard twitter API filtered with the language
parameter to harvest 9M Indonesian tweets from the period April--May
2020, covering the following topics: health, education, economy, and
government.  We discard threads that contain more than three
self-replies, and threads containing similar tweets (usually from
Twitter bots). Specifically, we discard a thread if 90\% of the tweets are
similar, as based on simple lexical overlap.\footnote{Two tweets are considered
  to be similar if they have a vocabulary overlap $\ge$80\%.}  We gather
threads that contain 3--5 tweets, and anonymize all mentions. This data
is used as the basis for the two discourse coherence tasks.
	
\begin{figure}[t!]
  \centering
  \includegraphics[width=6.3in]{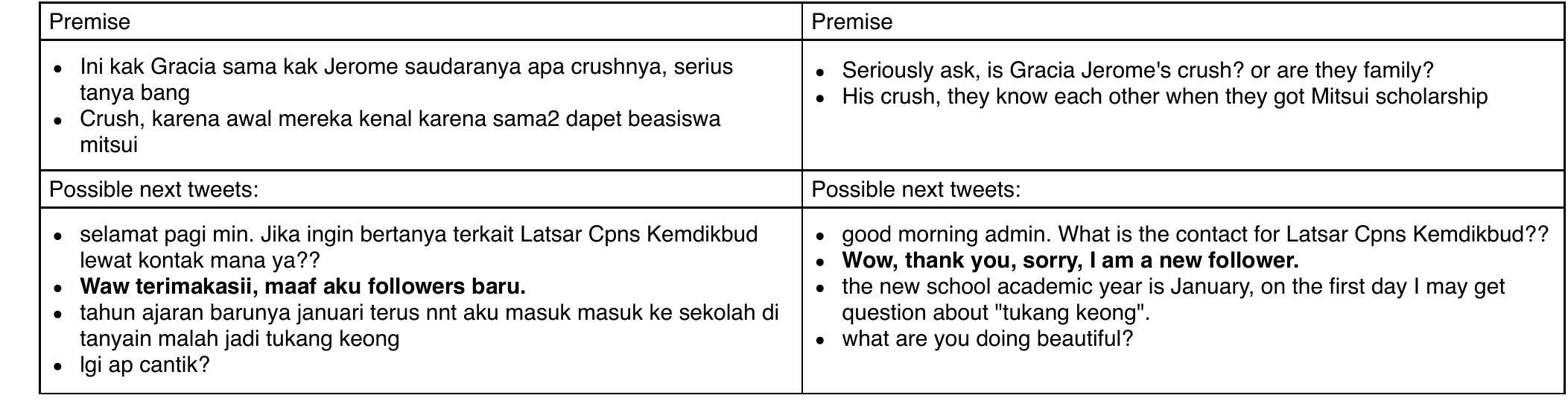}
  \caption{\label{fig:example} Example for the next tweet prediction
    task. To the left is the original Indonesian version and to the
    right is an English translation. The tweet indicated in bold is the correct next tweet.}
\end{figure}
	
\textbf{Next tweet prediction}.  To evaluate model coherence, we design
a next tweet prediction (NTP) task that is similar to the next sentence
prediction (NSP) task used to train BERT \cite{devlin2019bert}. In NTP,
each instance consists of a Twitter thread (2--4 tweets) that we call
the premise, and four possible options for the next tweet (see
\figref{example} for an example), one of which is the actual response
from the original thread.  In total, we construct 8,382 instances, where
the distractors are obtained by randomly picking three tweets from the
Twitter crawl. We ensure that there is no overlap between the next tweet
candidates in the training and test sets.
	
\textbf{Tweet ordering}.  For the second task, we propose a related but
more complex task of thread message ordering, based on the sentence
ordering task of \newcite{barzilay2008modeling} to assess text
relatedness. We construct the data by shuffling Twitter threads
(containing 3--5 tweets), and assessing the predicted ordering in terms
of rank correlation with the original. After removing all duplicates
messages, we obtain 7,608 instances for this task.

\section{Evaluation Methodology}
	
We provide details of the evaluation methodology in this section.
	
\textbf{Morpho-syntax/Sequence Labelling.} For POS tagging, we evaluate
by 5-fold cross validation using the partitions provided by
\newcite{kurniawan2018toward}. Unlike \newcite{kurniawan2018toward} who
use macro-averaged F1, we use the standard POS tag accuracy for
evaluation.  For NER, both corpora (NER UI and NER UGM) are from the
news domain. We convert them into \texttt{IOB2} format, and reserve 10\%
of the original training set as a validation set. We evaluate using
entity-level F1 over the provided test set.\footnote{We used the
  \texttt{seqeval} library to evaluate the POS and NER tasks.}  In
addition, we conducted our own in-house evaluation of the annotation
quality of both datasets by randomly picking 100 sentences and counting
the number of annotation errors. We found that NER UI has better
quality than NER UGM with 1\% vs.\ 30\% errors, respectively. Annotation
errors in NER UGM are largely due to low recall, i.e.\ annotating
named entities with the tag \texttt{O}.
	
For dependency parsing we do not apply 5-fold cross-validation for
UD-Indo-GSD, as it was released with a pre-defined test set, which
allows us to directly benchmark against previous work. UD-Indo-PUD, on
the other hand, only includes 1,000 sentences with no fixed test set, so
we evaluate via 5-fold cross-validation with fixed splits.\footnote{The
  split for cross validation is 70/10/20 for train/development/test,
  respectively. We first create 5 folds with non-overlapping test
  partitions, and for each fold we set the first portion of the
  remaining data as the development (and the rest as training data).}
Note that the text in UD-Indo-PUD was manually translated from documents
in other languages, while UD-Indo-GSD was sourced from texts authored in
Indonesian. Additionally, the translation quality of UD-Indo-PUD is low
in parts, which impacts on evaluation, as we return to discuss in
\secref{results}. We evaluate both dependency parsing datasets based on
the unlabelled attachment score (UAS) and labelled attachment score
(LAS).

\textbf{Semantics.}  Because the sentiment analysis data is low-resource
and imbalanced, we use stratified 5-fold cross-validation, and evaluate
based on F1 score.  For summarization, on the other hand, we use the
canonical splits provided by \newcite{kurniawan2018indosum}, and
evaluate the resulting summary with ROUGE (F1) \cite{lin2004rouge} in
the form of three different metrics: R1, R2, and RL.
	
	
	
\textbf{Discourse Coherence}. We do not perform 5-fold cross-validation over
NTP for two reasons. First, we need to ensure the distractors in the
test set do not overlap with the training or development sets, to avoid
possible bias because of dataset artefacts. Second, the size of the
dataset in terms of pair-wise labelling is actually four times the
reported size (\tabref{datastat}) as there are three distractors for
each thread. We evaluate the NTP task based on accuracy, meaning the
random baseline is 25\%.
	
For tweet ordering, we evaluate using Spearman's rank correlation
($\rho$). Specifically, we average the rank correlation between the gold
and predicted order of each thread in the dataset. 

\section{Comparative Evaluation}
	
\subsection{Baselines}
\label{sec:baseline}
	
Most of our experiments use a BiLSTM with 300d \fasttext \space
pre-trained Indonesian embeddings \cite{bojanowski2016enriching} as a
baseline.  Details of the baselines
are provided in \tabref{config}.

For extractive summarization baselines, we use the models of
\newcite{kurniawan2018indosum} and \newcite{cheng2016neural} as
baselines. \newcite{kurniawan2018indosum} propose a sentence tagging
approach based on a hidden Markov model, while \newcite{cheng2016neural}
use a hierarchical LSTM encoder with attention. In addition, we
present \model{Oracle} results, obtained by greedily maximizing the
ROUGE score between the reference summary and different combinations of
sentences from the document. \textsc{Oracle} denotes the upper bound for
the extractive summarization.
	
For next tweet prediction, we concatenate all premise tweets into a
single document, and use a BiLSTM and \fasttext word embeddings to
obtain the baseline document encoding. We structure this task as a
binary classification where we match the premise with each candidate
next tweet. We pick the tweet with the highest probability as the
prediction.  We use the same BiLSTM to encode the next tweet, and feed
the concatenated representations from the last hidden states into the
output layer.
	
For tweet ordering, we use a hierarchical BiLSTM model.  The first
BiLSTM is used to encode a single tweet by averaging all hidden
states. We use the second BiLSTM to learn the inter-tweet
ordering. We design the tweet ordering task as a sequence labelling
task, where we aim to obtain $P(r|t)$, the probability distribution
across rank positions $r$ for a given tweet $t$. Note that in this
experiment, each instance is comprised of 3--5 tweets, and we model the
task via multi-classification (with 5 classes/ranks). We perform
inference based on $P(r|t)$, where we decide the final rank based on the
highest sum of probabilities from the exhaustive enumeration of document
ranks.

\begin{table}[t!]
		\begin{center}
			\begin{adjustbox}{max width=1\linewidth}
				\begin{tabular}{lp{7cm}p{7cm}}
					\toprule
					\bf Task & \bf Baselines & \bf BERT models \\
					\midrule
					POS Tagging and NER & 
					\makecell[l]{\textbf{\newcite{lample-etal-2016-neural}}\footnote{The baseline code is available as \texttt{chars-lstm-lstm-crf} at
							\url{https://github.com/guillaumegenthial/tf_ner}.} \\
						A hierarchical BiLSTM + CRF with input:\\ 
					character-level embedding (updated),\\ 
					and word-level \fasttext embedding (fixed), \\
					lr: 0.001, epoch:100 with early stopping \\
					(patience = 5)} &  
					\makecell[l]{\bf Fine-tuning: \\
					adding a classification layer for each token,\\
					lr: 5e-5, epoch:100 with early stopping \\
					(patience = 5)	
					}\\
					\midrule
					
					Dependency parsing & 
					\makecell[l]{\textbf{1. \newcite{dozat2017deep}}, Bi-Affine \\ 
						parser, Embedding: \fasttext (fixed)\\ 
						\textbf{2. \newcite{rahman2020dense}}\sentfeat \\
						\textbf{3. \newcite{kondratyuk201975}}\sentfeat \\
						\textbf{4. \newcite{alfina2019gold}}\sentfeat 
					} &  
					\makecell[l]{\textbf{\newcite{dozat2017deep}}, Bi-Affine\\
					parser, Embedding: BERT output (fixed) 
					} \\
					\midrule
					
					Sentiment Analysis & 
					\makecell[l]{\textbf{1. 200-d BiLSTM} \\ 
						Embedding: \fasttext (fixed), \\
						lr: 0.001, epoch:100 with early stopping \\
						(patience = 5) \\
						\textbf{2. Naive Bayes and Logistic Regression}\\
						input: Byte-pair encoding (unigram+bigram)\footnote{We also experimented with simple term frequency, but observed lower performance so omit the results from the paper.}
					} &  
					\makecell[l]{\bf Fine-tuning: \\
						Input: 200 tokens; epoch: 20; lr: 5e-5; \\
						batch size: 30; warm-up: 10\% of the total steps; \\
						early stopping (patience = 5); \\
						Output layer uses the encoded [CLS]	
					} \\
					\midrule
					
					Summarization & 
					\makecell[l]{\textbf{1. \newcite{kurniawan2018indosum}\sentfeat} \\ 
						\textbf{2. \newcite{cheng2016neural}\sentfeat}	
					} &  
					\makecell[l]{ \bf \newcite{liu2019text}, extractive model,\\
						20,000 steps, lr: 2e-3, and tokens: 512.\footnote{ We checkpoint every 2,500 steps, and perform inference over the test set based on the top-3 best checkpoints according to the development set}
					} \\
					\midrule
					
					NTP & 
					\makecell[l]{\textbf{200-d BiLSTM (binary-class.)} \\ 
						Embedding: fastText (fixed), lr: 0.001,\\
						 epoch:100 with early stopping (patience = 20)
					} &  
					\makecell[l]{\bf Fine-tuning: \\
						Input: 60 tokens (for 1 single tweet); \\
						epoch: 20; learning rate; 5e-5; batch size: 20; \\
						warm-up: 10\% of the total steps; early stopping \\
						(patience = 5); Output layer uses the \\
						encoded [CLS]}\\
					\midrule
					
					Tweet Ordering & 
					\makecell[l]{\textbf{Hierarchical 200-d BiLSTMs (multi-class.)} \\ 
						Embedding: fastText (fixed), lr: 0.001,\\
						epoch:100 with early stopping (patience = 20)
					} &  
					\makecell[l]{\bf Fine-tuning: \\
						Input: 50 tokens (for 1 single tweet); \\
						epoch: 20; learning rate; 5e-5; batch size: 20; \\
						warm-up: 10\% of the total steps; early stopping \\
						(patience = 5); BERT
                                  fine-tuning is based on the \\
						\newcite{liu2019text} trick (alternated seq.)}
					\\
					
					\bottomrule
				\end{tabular}
			\end{adjustbox}
		\end{center}
		\caption{Comparison of baselines and BERT-based models
                  for all \indolem tasks. All listed models were
                  implemented and run by the authors, except for those
                  marked with ``\sentfeat'' where the results are
                  sourced from the original paper.}
		\label{tab:config}
\end{table}
	
\subsection{BERT Benchmarks}
	
To benchmark \indobert, we compare against two pre-existing BERT models:
multilingual BERT (``\mbert''), and a monolingual BERT for Malay
(``\malaybert'').\footnote{\url{https://huggingface.co/huseinzol05/bert-base-bahasa-cased}}
\mbert is trained by concatenating Wikipedia documents for 104 languages
including Indonesian, and has been shown to be effective for zero-shot
multilingual tasks \cite{wu2019beto,wang2019cross}. \malaybert is a
a publicly available model that was trained on Malay documents from
Wikipedia, local news sources, social media, and some translations from
English. We expect \malaybert to provide better representations than
\mbert for the Indonesian language, because Malay and Indonesian
are mutually intelligible, with many lexical similarities, but
noticeable differences in grammar, pronunciation and vocabulary.
	
	
For the sequence labelling tasks (POS tagging and NER), sentiment analysis, NTP, and tweet ordering task, the fine-tuning procedure is detailed in \tabref{config}.

For dependency parsing, we follow \newcite{nguyen2020phobert} in
incorporating BERT into the BiAffine dependency parser
\cite{dozat2017deep} by replacing the word embeddings with the
corresponding contextualized representations. Specifically, we generate
the BERT embedding of the first WordPiece token as the word embedding,
and train the BiAffine parser in its default configuration. In addition,
we also benchmark against a pre-existing fine-tuned version of \mbert
trained over 75 concatenated UD datasets in different languages
\cite{kondratyuk201975}.
	
	
For summarization, we follow \newcite{liu2019text} in encoding the
document by inserting the tokens [CLS] and [SEP] between sentences. We
also apply alternating segment embeddings based on whether the position
of a sentence is odd or even. On top of the pre-trained model, we use a
second transformer encoder to learn inter-sentential relationships. The
input is the encoded [CLS] representation, and the output is the
extractive label $y\in\{0,1\}$ (1 = include in summary; 0 = don't
include).
	
%
	
\section{Results}
\label{sec:results}
	
\tabref{results_seqlab} shows the results for POS tagging and
NER. \mbert, \malaybert, and \indobert perform very similarly over the
POS tagging task, well above the BiLSTM baseline. This indicates that
all three contextual embedding models are able to generalize well over
low-level morpho-syntactic tasks. Given that Indonesian and Malay share
a large number of words, it is not surprising that \malaybert performs
on par with \indobert for POS tagging. On the NER tasks, both \malaybert
and \indobert outperform \mbert, which performs similarly to or slightly
above the BiLSTM. This is despite \mbert having been trained on a much
larger corpus, and having seen many more entities during
training. \indobert slightly outperforms \malaybert.
	
In \tabref{res_dep}, we show that augmenting the BiAffine parser with
the pre-trained models yields a strong result for dependency parsing,
well above previously-published results over the respective
datasets. Over UD-Indo-GSD, \indobert outperforms all methods on
both metrics. The universal fine-tuning approach \cite{kondratyuk201975}
yields similar performance as BiAffine + \fasttext, while augmenting
BiAffine with \mbert and \malaybert yields lower UAS and LAS scores than
\indobert. Over UD-Indo-PUD, we see that augmenting BiAffine
with \mbert outperforms all methods including \indobert. Note that
\newcite{kondratyuk201975} is trained on the original version of
UD-Indo-PUD, and \newcite{alfina2019gold} is based on 10-fold
cross-validation, meaning the results are not 100\% comparable.
	
To better understand why \mbert performs so well over UD-Indo-PUD, we
randomly selected 100 instances for manual analysis. We found that 44
out of the 100 sentences contained direct borrowings of foreign words
(29 names, 10 locations, and 15 organisations), some of which we would
expect to be localized into Indonesian, such as: \lex{St.\ Rastislav},
\lex{Star Reach}, \lex{Royal National Park Australia}, and \lex{Zettel's
  Traum}. We also thoroughly examined the translation quality and found
that roughly 20\% of the sentences are low-quality translations. For
instance, \lex{Ketidaksesuaian data ekonomi dan retorika politik tidak
  asing, atau seharusnya tidak asing} is not a natural sentence in
Indonesian.

For the semantic tasks, \indobert outperforms all other methods for both
sentiment analysis and extractive summarization
(\tabref{result_semantic}). For sentiment analysis, the improvement over
the baselines is impressive: +13.2 points over naive Bayes, and +7.5
points over \mbert. As expected, \malaybert also performs well for
sentiment analysis, but substantially lower than \indobert. For
summarization, \mbert and \malaybert achieve similar performance, and only
outperform \newcite{cheng2016neural} by around 0.5 ROUGE
points. \indobert, on the other hand, is 1--2 ROUGE points better.
	
	\begin{table}[t!]
		\begin{center}
            \begin{adjustbox}{max width=0.8\linewidth}
				\begin{tabular}{L{6cm}ccc}
					\toprule 
					\multirow{2}{*}{\textbf{Method}} & \textbf{POS tagging} & \textbf{NER UGM} & \textbf{NER UI} \\
					& \textbf{Acc} & \textbf{F1} & \textbf{F1} \\
					\midrule 
					BiLSTM-CRF~\cite{lample-etal-2016-neural} & 95.4 & 70.9 & 82.2 \\
					\mbert & \textbf{96.8} & 71.6 & 82.2 \\
					\malaybert  & \textbf{96.8} & 73.2 & 87.4 \\
					\indobert & \textbf{96.8} & \textbf{74.9} & \textbf{90.1} \\
					\bottomrule
				\end{tabular}
			\end{adjustbox}
		\end{center}
		\caption{\label{tab:results_seqlab} Results on POS and NER tasks
			using accuracy averaged over five folds for POS tagging task,
			and entity-level F1 over the test set for the NER tasks.}
	\end{table}
	
	\begin{table}[t!]
		\begin{center}
			\begin{adjustbox}{max width=1\linewidth}
				\begin{tabular}{ll}
					\begin{tabular}{lC{1.2cm}C{1.2cm}}
						\toprule
						\multirow{2}{*}{\textbf{Method}} & \multicolumn{2}{c}{\textbf{UD-Indo-GSD}} \\
						& \textbf{UAS} & \textbf{LAS} \\
						\midrule
						\newcite{rahman2020dense}* & 82.56 & 76.04 \\
						\newcite{kondratyuk201975} & 86.45	& 80.10 \\
						BiAffine w/ \fasttext & 85.25 & 80.35\\
						BiAffine w/ \mbert & 86.85 & 81.78 \\
						BiAffine w/ \malaybert & 86.99 & 81.87 \\
						BiAffine w/ \indobert & \textbf{87.12} & \textbf{82.32}\\
						\bottomrule
					\end{tabular}
					&
					\begin{tabular}{lC{1.2cm}C{1.2cm}}
						\toprule
						\multirow{2}{*}{\textbf{Method}} & \multicolumn{2}{c}{\textbf{UD-Indo-PUD}} \\
						& \textbf{UAS} & \textbf{LAS} \\
						\midrule
						\newcite{alfina2019gold}* & 83.33 & 79.39 \\
						\newcite{kondratyuk201975}* & 77.47 & 56.90 \\
						BiAffine w/ \fasttext & 84.04 &79.01 \\
						BiAffine w/ \mbert & \textbf{90.58} & \textbf{85.44} \\
						BiAffine w/ \malaybert  & 88.91 & 83.56 \\
						BiAffine w/ \indobert & 89.23 & 83.95 \\
						\bottomrule
					\end{tabular}
				\end{tabular}
			\end{adjustbox}
		\end{center}
		\caption{\label{tab:res_dep} Results for dependency
			parsing. Methods marked with `*' (from previous work) do not use the same test partition.}
	\end{table}

	\begin{table}[!t]
		\begin{center}
			\begin{adjustbox}{max width=1\linewidth}
				\begin{tabular}{lll}
					\begin{tabular}{L{4cm}c}
						\toprule 
						\multirow{2}{*}{\textbf{Method}} & \textbf{Sentiment}\\
						& \textbf{Analysis (F1)} \\
						\midrule
						Naive Bayes & 70.95 \\
						Logistic Regression & 72.14 \\
						BiLSTM w/ \fasttext & 71.62 \\
						\mbert & 76.58 \\
						\malaybert & 82.02 \\
						\indobert & \textbf{84.13} \\
						\bottomrule
					\end{tabular}
					& 
					\begin{tabular}{lC{1cm}C{1cm}C{1cm}}
						\toprule 
						\multirow{2}{*}{\textbf{Method}} & \multicolumn{3}{c}{\textbf{Summarization (F1)}} \\
						& \textbf{R1} & \textbf{R2} & \textbf{RL} \\
						\midrule
						\textsc{Oracle} & 79.27 & 72.52 & 78.82 \\
						\newcite{kurniawan2018indosum} & 17.62 & 4.70 & 15.89 \\
						\newcite{cheng2016neural} & 67.96 & 61.65 & 67.24 \\
						\mbert & 68.40 & 61.66 & 67.67 \\
						\malaybert & 68.44 & 61.38 & 67.71 \\
						\indobert & \textbf{69.93} & \textbf{62.86} & \textbf{69.21} \\
						\bottomrule
					\end{tabular}
				\end{tabular}
			\end{adjustbox}
		\end{center}
		\caption{\label{tab:result_semantic} Results over the semantic tasks.}
	\end{table}

	\begin{table}[h!]
		\begin{center}
            \begin{adjustbox}{max width=0.65\linewidth}
				\begin{tabular}{L{4cm}cc}
					\toprule 
					\multirow{2}{*}{\textbf{Method}} & \textbf{Next Tweet} & \textbf{Tweet Ordering} \\
					& \textbf{Prediction (Acc)} & \textbf{($\rho$)} \\
					\midrule 
					Random & 25.0 & 0.00 \\
					Human (100 samples) & 90.0 & 0.61 \\
					BiLSTM w/ \fasttext & 73.6 & 0.45 \\
					\mbert & 92.4 & 0.53 \\
					\malaybert  & 93.1 & 0.51 \\
					\indobert & \textbf{93.7} & \textbf{0.59} \\
					\bottomrule
				\end{tabular}
			\end{adjustbox}
		\end{center}
		\caption{\label{tab:result_coherence} Results for discourse
			coherence. ``Human'' is the oracle performance by a human
			annotator.}
\end{table}
	
Lastly, in \tabref{result_coherence}, we observe that \indobert is once
again substantially better than the other models at discourse coherence
modelling, despite its training not including next sentence prediction
(as per the English BERT). To assess the difficulty of the NTP task, we
randomly selected 100 test instances, and the first author (a native
speaker of Indonesian) manually predicted the next tweet. The human
performance was 90\%, lower than the pre-trained language models. For
the tweet ordering task, we also assess human performance by randomly
selecting 100 test instances, and found the rank correlation score of
$\rho = 0.61$ to be slightly higher than \indobert. The gap between
\indobert and the other BERT models was bigger on this task.
	
Overall, with the possible exception of POS tagging and NTP, there is
substantial room for improvement across all tasks, and our hope is
that \indolem can serve as a benchmark dataset to track progress in
Indonesian NLP.
	
\section{Conclusion}
	
In this paper, we introduced \indolem, a comprehensive dataset
encompassing seven tasks, spanning morpho-syntax, semantics, and
discourse coherence. We also detailed \indobert, a new BERT-style
monolingual pre-trained language model for Indonesian. We used \indolem
to benchmark \indobert (including comparative evaluation against a broad
range of baselines and competitor BERT models), and showed it to achieve
state-of-the-art performance over the dataset.
	
	%
	%

\section*{Acknowledgements}

We are grateful to the anonymous reviewers for their helpful feedback
and suggestions. The first author is supported by the Australia Awards
Scholarship (AAS), funded by the Department of Foreign Affairs and Trade
(DFAT), Australia.  This research was undertaken using the LIEF
HPC-GPGPU Facility hosted at The University of Melbourne. This facility
was established with the assistance of LIEF Grant LE170100200.
	
	\bibliographystyle{coling}
	\bibliography{coling2020}

\end{document}